\documentclass[5p,times]{elsarticle}

\usepackage[utf8]{inputenc}

\graphicspath{{./figures/}{./resources/}}
\DeclareGraphicsExtensions{.pdf,.png,.jpg,.jpeg}

\usepackage{amssymb}
\usepackage{amsmath}
\usepackage{mdwmath}
\usepackage{mdwtab}

\usepackage{tabularx, booktabs, makecell, multirow}
\newcommand{\ra}[1]{\renewcommand{\arraystretch}{#1}}
\usepackage[flushleft]{threeparttable}
\usepackage{xcolor}

\usepackage{wrapfig}
\usepackage{tikz}
\usetikzlibrary{shapes,arrows}
\usepackage{pgfplots}
\pgfplotsset{compat=1.9}

\usepackage{mathtools}

\usepackage{siunitx}

\usepackage{tikz}
\usetikzlibrary{shapes, arrows, decorations, positioning, calc, shadows, trees, mindmap}
\usepackage{listings}

\usepackage[figuresright]{rotating}
\usepackage{afterpage}
\usepackage{comment}
\usepackage{xspace}
\makeatletter
\DeclareRobustCommand\onedot{\futurelet\@let@token\@onedot}
\def\@onedot{\ifx\@let@token.\else.\null\fi\xspace}

\def\ie{\emph{i.e}\onedot}

\makeatother

\usepackage{hyperref}

\definecolor{codegreen}{rgb}{0,0.6,0}
\definecolor{codegray}{rgb}{0.5,0.5,0.5}
\definecolor{codepurple}{rgb}{0.58,0,0.82}
\definecolor{backcolour}{rgb}{0.95,0.95,0.92}

\lstdefinestyle{mystyle}{
  backgroundcolor=\color{backcolour}, commentstyle=\color{codegreen},
  keywordstyle=\color{magenta},
  numberstyle=\tiny\color{codegray},
  stringstyle=\color{codepurple},
  basicstyle=\ttfamily\footnotesize,
  breakatwhitespace=false,         
  breaklines=true,                 
  captionpos=b,                    
  keepspaces=true,                 
  numbers=left,                    
  numbersep=5pt,                  
  showspaces=false,                
  showstringspaces=false,
  showtabs=false,                  
  tabsize=2
}

\usepackage{footnote}
\makesavenoteenv{tabular}
\makesavenoteenv{table}

\usepackage[nolist]{acronym}% http://ctan.org/pkg/acronym

\begin{acronym}[MACHU]
  \acro{ml}[ML]{Machine Learning}
  \acro{ai}[AI]{Artificial Intelligence}
  \acro{mlops}[MLOps]{Machine Learning Operation}
  \acro{mse}[MSE]{mean squared error}
  \acro{mpe}[MPE]{mean percentage error}
  \acro{shap}[SHAP]{SHapley Additive exPlanations}
  \acro{pdf}[PDF]{Probability Distribution Function}
  \acro{ems}[EMS]{energy management system}
  \acroplural{ems}[EMSes]{energy management systems}
  \acro{sm}[SM]{smart meter}
  \acro{ied}[IED]{intelligent electronic device}
  \acro{wams}[WAMS]{wide-area measurement system}
  \acroplural{wams}[WAMS]{wide-area measurement system}
  \acro{lv}[LV]{low voltage}
  \acro{mv}[MV]{medium voltage}
  \acro{hv}[HV]{high voltage}
  \acro{dso}[DSO]{distribution system operator}
  \acro{ied}[IED]{intelligent electronic device}
  \acro{nilm}[NILM]{nonintrusive load monitoring}
  \acro{pv}[PV]{photovoltaic}
  \acro{ev}[EV]{electric vehicle}
  \acro{mape}[MAPE]{mean absolute percentage error}
  \acro{smape}[SMAPE]{symmetric mean absolute percentage error}
  \acro{xgb}[XGB]{extreme gradient boosting}
  \acro{lr}[LR]{linear regression}
  \acro{mlp}[MLP]{multi-layer perceptrons}

  \acro{sargon}[SARGON]{smart energy domain ontology}
  \acro{saref}[SAREF]{smart applications reference}
  \acro{foaf}[FOAF]{acronym of friend of a friend}
  \acro{owl}[OWL]{web ontology language}
  \acro{pca}[PCA]{principal component analysis}
\end{acronym}

\journal{Expert Systems with Applications}

\begin{document}

\begin{frontmatter}

\title{Data Model Design for Explainable Machine Learning-based Electricity Applications}

%\title{Data Model Design for Explainable Electricity Applications}

%\title{Data Model Design for Electricity Applications: an Explainable Household Forecasting Case Study}
%\title{Data Model Design for Explainable Electricity Applications}
%\title{Data and Model Design for Explainable  Household Electricity Forecasting}

%\author{Carolina~Fortuna, Gregor~Cerar, Blaž~Bertalanič,  Andrej~Čampa, Mihael~Mohorčič}

\author[ijs]{Carolina~Fortuna\corref{cor}}
\ead{carolina.fortuna@ijs.si}

\author[ijs]{Gregor~Cerar}
\ead{gregor.cerar@ijs.si}

\author[ijs]{Bla\v{z}~Bertalani\v{c}}
\ead{blaz.bertalanic@ijs.si}

\author[ijs]{Andrej~\v{C}ampa}
\ead{andrej.campa@ijs.si}

\author[ijs]{Mihael~Mohor\v{c}i\v{c}}
\ead{miha.mohorcic@ijs.si}

\affiliation[ijs]{organization={Jožef Stefan Institute},
    %city={Ljubljana},
    country={Slovenia}
}

%\author[label1]{Gregor~Cerar\corref{cor1}}
%\ead{gregor.cerar@ijs.si}

% \author[label1]{Blaž~Bertalanič\corref{cor1}}
% \ead{blaz.bertalanic@ijs.si}

% \author[label1]{Andrej~Čampa}
% \ead{andrej.campa@ijs.si}

% \author[label1]{Carolina~Fortuna}
% \ead{carolina.fortuna@ijs.si}

% \affiliation[label1]{%
%     organization={Department of Communication Systems, Jožef~Stefan~Institute},%
%     addressline={Jamova~cesta~39},%
%     city={Ljubljana},%
%     country={Slovenia}%
% }

%\cortext[cor1]{These authors contributed equally}

%\maketitle

\begin{abstract}
%What is important?
The transition from traditional power grids to smart grids,
significant increase in the use of renewable energy sources,
and soaring electricity prices has triggered a digital transformation of the energy infrastructure that enables new, data driven, applications often supported by machine learning models. 
%what is missing?
However, the majority of the developed machine learning models rely on univariate data. To date,  a structured study considering the role meta-data and additional measurements resulting in multivariate data is missing.
%How do we solve it?
In this paper we propose a taxonomy that identifies and
structures various types of data related to energy applications. The taxonomy can be used to guide application specific data model development for training machine learning models. Focusing on a household electricity forecasting application, we validate the effectiveness of the proposed taxonomy in guiding the selection of the features for various types of models. As such, we study of the effect of domain, contextual and behavioral features on the forecasting accuracy of four interpretable machine learning techniques and three openly available datasets. Finally, using a feature importance techniques, we explain individual feature contributions to the forecasting accuracy.

\end{abstract}



%\item Taxonomy for guiding energy data model design
%\item Suitable data models and corresponding features improve energy consumption forecast
%\item Automated feature management decreases the data preparation complexity and cost
%\item Emerging feature stores improve the speed of data processing and enrichment


%\begin{IEEEkeywords}
%data model, model design, electricity forecast, explainability
%\end{IEEEkeywords}

%\begin{keywords}
%A \sep B
%\end{keywords}

\end{frontmatter}

\section{Introduction}
\label{sec:intro}

% What is important? (2 paragraphs)
The transition from traditional power grids to smart grids, significant increase in the use of renewable energy sources, and soaring electricity prices has led to an increase in complexity~\cite{fortuna2022}, particularly with the adoption of \acp{sm}, \acp{ems}, and \acp{ied} at the \ac{lv} level. These devices enable innovative energy~\cite{dileep2020survey} and non-energy applications~\cite{chuang2020monitoring, cerar2023modelSelection}, such as energy cost optimization and matching consumption with self-production from renewable energy sources. On the \ac{dso} side of the \ac{lv} grid, reliability and latency are the main challenges, and complete observability of the \ac{lv} grid for each substation is crucial. While \acp{wams} already monitor and collect data at the \ac{mv} and \ac{hv} levels, the data is mainly used for observability and critical situation handling. However, the data collected in the \ac{lv} grid can be processed and used to enrich the collected data at the \ac{mv} and \ac{hv} levels, creating a limited control loop that extends from the observability of the \ac{hv} grid to the prosumer.

The increased penetration of \acp{ied}, \acp{wams}, and \acp{ems} generates large amounts of data, leading to the adoption of big data and machine learning techniques and opens opportunities for energy consumption forecasting integrated into applications serving all grid segments~\cite{wang2018review, 8937810}. A taxonomy of such applications including energy saving, appliance recognition, occupancy detection, preference detection as well as a taxonomy of related datasets and techniques supporting  applications have been surveyed in~\cite{HIMEUR2020110404}. The study concluded that the datasets 1) are generally limited geographically, in duration, type of appliances or sampling frequency, and 2) do not capture exogenous conditions, such as the weather temperature, humidity as well as meta-data such as building insulation, nature of buildings along with other data specifying the number of individuals, job quality and age.

% What is missing? (1-2 short sentences) - What improvements can happen in that area?

Furthermore, using only direct instantaneous measurements may be insufficient to accurately forecast energy consumption\cite{al2022buildings,HIMEUR2020110404}. Next, the authors of~\cite{al2022buildings} found that meta-data such as building height, wall area and orientation were employed in a number of studies revealing that such features affect the prediction for target variables such as heating and cooling load. The studies surveyed largely relied on traditional machine learning techniques as has also been noted in another recent survey~\cite{nti2020electricity}. The main advantages of such methods are that they are inherently interpretable~\cite{BAUR2024100358} and tend to have lower computation complexity compared to their deep learning counterparts.

The majority of forecasting studies employ univariate data consisting of aggregated household consumption collected from a main \ac{ied}, and develop models that learn the likely distribution of the values in the metering data to predict future energy consumption~\cite{kong2017short}. Sometimes lagged values are also used in the model development~\cite{AMJADY200912281}. A relatively lower number of studies, such as ~\cite{zhang2018forecasting}, estimated consumption using timestamp (month, hour, week of year, day of week, season), weather (condition, severity, temperature, humidity), and energy price. In~\cite{Lim2020DeepLearning-BasedAnalysis}, the ML-based estimation was done using a timestamp, electricity contract type, energy consumption, and city area. However, in such works, the features used to train the model are developed by machine learning  experts and are less intuitive for decision makers or energy domain practitioners~\cite{10.1145/3544903.3544905}. 
While taxonomies and semantic modeling efforts for representing various domains~\cite{bertoli2022semantic}, including technical aspects of energy infrastructure~\cite{daniele2015created}, have been developed, a structured study considering the role meta-data and additional measurements in guiding model selection, accuracy and explainability of electricity applications to complement existing works and guide towards more comprehensive models is missing.  

% How do we solve it? (the contributions paragraphs and bullet points)
In this paper we propose a taxonomy that identifies and structures various types of data related to electricity applications. Based on this taxonomy, data models can be designed and implemented in database-like systems or feature stores for machine learning model training. Based on the proposed taxonomy, we identified and employed relevant data sources related to the energy domain (i.e. domain specific features) as well as contextual and behavioural. Together with selected interpretable machine learning models, we study the contribution and importance of each feature category on the final performance of the models for household electricity forecasting while also reflecting on individual feature influence using a feature importance based explainability technique~\cite{BAUR2024100358}.

The contributions of this paper are as follows:
\begin{itemize}
    \item The proposed taxonomy and the validations of its effectiveness its effectiveness on a household electricity forecasting problem.
    \item The study of the effect of domain, contextual and behavioral features on the forecasting accuracy of various categories of model developed with four traditional but interpretable machine learning techniques and three openly available datasets. We find that feature engineering can improve the performance of the model by $\approx$3.73, $\approx$4.70 and $\approx$4.30 percentage points in terms of \ac{mpe} metric, compared to just utilising raw time series on three relevant datasets. 
    \item Feature importance analysis reveals that domain-specific features contribute the most to the final performance with contributions between $65.5\%$ to $83.1\%$,  contextual features contribute between $10.8\%$ and $23.6\%$ while behavioral features contribute between $6.1\%$ and $13.2\%$. This analysis also reveals that more sub-metering data can help extract better behavioral patterns from households, which can have a high impact on the forecasting accuracy of future models. 
\end{itemize}

The rest of the paper is structured as follows. Section~\ref{sec:rw} summarizes the related work, Section~\ref{sec:taxonomy} elaborates on the proposed taxonomy for designing energy data models, Section~\ref{sec:methodology} discusses methodology and experiment details while Section~\ref{sec:evaluation} details the evaluation, where we examine contribution of feature groups, and individual features. Finally, Section~\ref{sec:conclusions} concludes the paper.

\section{Related work}
\label{sec:rw}
%A number of taxonomies have been proposed for various aspects of the energy domain with focus on electricity. 
We identify a) taxonomies developed manually by surveying published works and b) taxonomies developed automatically by extracting information from data. 

Focusing on the manually developed taxonomies, the authors of~\cite{HIMEUR2020110404} developed a taxonomy of existing domestic power consumption datasets and their suitability to enable energy specific applications. They identified appliance level and aggregated level datasets. The appliance level datasets enable energy saving, appliance recognition, occupancy detection, preference detection and anomaly detection applications. The aggregated level datasets enable energy saving, energy dissagregation and demand forecasting application. A 3-tier taxonomy of machine learning applications for virtual power plants has been developed in~\cite{SIERLA2022104174} while a taxonomy of machine learning in smart buildings focused on occupant-centric and device-centric techniques has been developed in~\cite{10.1145/3311950}.  The top tier categories under which current research works fall are optimization, forecasting and classification. The authors of~\cite{CHARBONNIER2022119188} developed a taxonomy of coordination techniques for the edge of the electricity grid. The taxonomy distinguished between direct and indirect control in the first level, with the indirect control being further split into mediated, bilateral and implicit coordination. Perhaps, conceptually the most relevant manually developed taxonomy for the work in this paper is the one developed for promoting and arguing for the need for interpretable features~\cite{10.1145/3544903.3544905}. Depending on the users, \ie decision makers vs. theorists and developers, the taxonomy guides the most suitable types of feature spaces and transforms to be used. They identify properties related to model-ready feature spaces, interpretable feature spaces, as well as properties related to all feature spaces. In this work, we complement this work by developing a context based feature taxonomy.

A data driven taxonomy of electrical appliances, automatically extracted from data, has been developed by clustering appliance load signatures~\cite{4266955}. A framework to map segments of disaggregated electricity consumption to daily activities has been proposed in~\cite{AHMADIKARVIGH2016337}. The activities and relationships are part of a taxonomy that is then mapped to an ontology.

While not guided by a taxonomy, a number of works focused on feature selection for electricity applications. For instance, in~\cite{SADEGHIANPOURHAMAMI201798} a number of domain specific features, such as measured real power, reactive power and apparent power, as well as other synthetic or derivatives extracted from the measurements such as root mean square, slope and coefficients of transforms such as wavelet. They investigate the model performance as a function of various feature subsets with a focus on appliance classification for  non-intrusive load monitoring. Another recent example, investigated the relative importance of variable groups such as domain specific measurements, as well as contextual such as economic, weather and financial metrics for an electricity consumption application~\cite{ALBUQUERQUE2022115917}. Furthermore, several studies, such as ~\cite{Spichakova2019FeatureEngineeringForShort-TermForecastOfEnergyConsumption, Adika2012NorthAmericanPowerSymposium(NAPS), parkinson2021overcooling} show that electricity consumption is influenced by the behavior of the inhabitants. These are, for instance, personal hygiene, person's age, person's origins, work habits, cooking habits, social activities~\cite{Spichakova2019FeatureEngineeringForShort-TermForecastOfEnergyConsumption}, and wealth class~\cite{Adika2012NorthAmericanPowerSymposium(NAPS)}. People of different ages have diverse electricity consumption patterns because of the difference in sleep habits and lifestyles in general. Furthermore, the gender of the inhabitants also plays a role as men and women have different preferences when it comes to air temperature, water temperature~\cite{parkinson2021overcooling} and daily routines. While such data is useful for estimating and optimizing energy consumption, it raises ethical and privacy concerns as well. 

The most intuitive way to start building features from raw energy time series signals is to search for their statistical properties as employed in~\cite{SADEGHIANPOURHAMAMI201798}. Such features are in general easy to compute and interpret but tend to raise the number of training variables. Among others,~\cite{yu2015towards, Fagaras_9658990} used such features to forecast energy usage in smart grids and residential area,~\cite{Chowdhury_9249667} employed them to detect appliances for non intrusive load monitoring, while~\cite{Ouyang2017} proposed the use of statistical features for power consumption anomaly detection.

However, increasing the number of input variables through such techniques can also increase computational complexity. To balance the performance/complexity trade-offs, dimensionality reduction techniques help reduce the number of input features while keeping as much variation as possible. %Dimensionality reduction techniques are widely used in energy domain for solving different set of problems. 
\Ac{pca} is one of the most used techniques and has been utilised in forecasting of energy production~\cite{ZHANG2019180, ge2020hybrid}, energy consumption estimation~\cite{PLATON201510}, and \ac{nilm} disaggregation and appliance classification~\cite{machlev2020dimension, moradzadeh2020improving}. While the \ac{pca} dimensionality reduction requires more effort to  interpret the effects of specific input data to the final performance of the model, the more recent deep learning based representation learning methods such as autoencoder based ~\cite{Kaur_9473748} are far less interpretable. However, they reduce the dimension of input data and report good performance for various applications including energy production forecast~\cite{Kaur_9473748}, electricity price forecast~\cite{POURDARYAEI2024121207}.

Competitions such as GEPIII~\cite{miller2021limitations} and the Post-COVID energy consumption forecasting~\cite{mostafa2022covid}, involving thousands of participants, have offered valuable insights into the field of energy consumption forecasting, shedding light on the accomplishments of leading participants. These solutions have empirically found the best combination of preprocessing, feature selection, model types, and post-processing strategies. The findings showed that large ensembles of mainly gradient boosting trees with significant preprocessing of the training data were found to be the best solutions for this application~\cite{miller2021limitations}.

To complement the existing works and promote more systematic, informed and explainable feature development for training machine learning models for electricity applications we propose a new taxonomy of relevant features distinguishing between domain specific, contextual and behavioral on its first level.

\section{Taxonomy for Guiding Data Model Design}
\label{sec:taxonomy}

\begin{figure*}[htbp]
\centering
\resizebox{0.7\linewidth}{!}{
\begin{tikzpicture}[
	scale=0.9,
	grow cyclic,
	edge from parent/.style = {draw, -, thick, red},
	%every node/.style = {font=\footnotesize},
	sloped,
	box/.style = { shape=rectangle, rounded corners, draw=red },
	level/.style = { align=center },
	level 0/.style = { level, font=\bfseries\normalsize, align=center, ultra thick },
	level 1/.style = { level, level distance=8em, font=\bfseries\small, sibling angle=120 },
	level 2/.style = { level, level distance=8em, font=\bfseries\small, sibling angle=30 },
	level 3/.style = { level, level distance=8em, font=\footnotesize, sibling angle=22 },
	level 4/.style = { level, level distance=7em, font=\scriptsize, sibling angle=18 }
]
\node [level 0, box] {Applicable\\Features}
	child [level 1] { node [box] {Domain\\Specific\\Features}
		child [level 2, sibling angle=40, rotate=10] { node {PV power plant\\measurements}
		    child [level 4, sibling angle=16] { node{batteries} }
			child [level 4, sibling angle=16] { node{voltage} }
			child [level 4, sibling angle=16] { node{current} }
			child [level 4, sibling angle=16] { node{power} }
			child [level 4, sibling angle=16] { node{energy} }
		}
		child [level 2, sibling angle=40, rotate=-3] { node {Electric\\vehicles}
		    child [level 4, rotate=11, sibling angle=18] { node{capacity} }
		    child [level 4, rotate=11, sibling angle=18] { node{charge rate} }
		    child [level 4, rotate=11, sibling angle=18] { node{discharge rate} }
		}
		child [level 2, sibling angle=40, rotate=6, level distance=9em] { node {Household\\Measurements}
			child [level 4, rotate=14, sibling angle=22] { node{Appliances} }
			child [level 4, rotate=14, sibling angle=22] { node{Active power} }
			child [level 4, rotate=14, sibling angle=22] { node{Voltage} }
			child [level 4, rotate=14, sibling angle=22] { node{Current} }
			child [level 4, rotate=14, sibling angle=22] { node{Reactive Power} }
			child [level 4, rotate=14, sibling angle=22] { node{Apparent power} }
			child [level 4, rotate=11, sibling angle=22] { node{Phase} }
		}
		child [level 2,sibling angle=40, rotate=20] { node {wind power plant\\measurements}
			child [level 4] { node {Wind capacity}}
			child [level 4] { node {Wind generation}}
		}
	}
	child [level 1] { node [box] {Contextual\\Features}
		child [level 2, level distance=5em, rotate=-10] { node {weather\\conditions}
			child [level 3, rotate=-30] { node {relative\\humidity}}
			child [level 3, rotate=-30] { node {temperature}}
			child [level 3, rotate=-30] { node {pressure}}
			child [level 3, rotate=-30] { node {wind}
				child [level 4, sibling angle=16, rotate=10] { node {direction}}
				child [level 4, sibling angle=16, rotate=15] { node {speed}}
			}
			child [level 3, rotate=-30] { node {visibility}}
			child [level 3, rotate=-30] { node {precipitation}}
			child [level 3, rotate=-30, level distance=8em] { node {cloud coverage}}
		}
		child [level 2, rotate=15] { node {building\\properties}
			child [level 4, sibling angle=16, rotate=-10] { node {plug load}}
			child [level 4, sibling angle=16, rotate=-10] { node {age}}
			child [level 4, sibling angle=16, rotate=-10] { node {type}}
			child [level 4, sibling angle=16, rotate=-10] { node {area density}}
			child [level 4, sibling angle=16, rotate=-10] { node {orientation}}
			child [level 4, sibling angle=16, rotate=-10] { node {geolocation}}
		}
		child [level 2, rotate=30, level distance=4em] { node {time}
			child [level 3] { node {daytime duration}}
			child [level 3] { node {time of the day}}
			child [level 3] { node {seasons}}
		}
		child [level 2, rotate=60, level distance=4em] { node {geolocation}
			child [level 3, rotate=0, level distance=4em] { node {latitute}}
			child [level 3, rotate=0, level distance=5em] { node {longitude}}
			child [level 3, rotate=0, level distance=4em] { node {region}}
		}
	}
	child [level 1] { node [box] {Behavioral\\Features}
		child [level 2] { node {wealth\\class}}
		child [level 2] { node {social\\activities}
			child [level 4, level distance=6em, rotate=-40] { node {weekday}}
			child [level 4, level distance=6em, rotate=-40] { node {weekend}}
			child [level 4, level distance=6em, rotate=-40] { node {holidays}}
			child [level 4, level distance=7em, rotate=-40] { node {near-holidays}}
		}
		child [level 2] { node {heating}}
		child [level 2] { node {age}}
		child [level 2] { node {work\\schedule}}
		child [level 2] { node {personal\\hygiene}
			child [level 3] { node {duration}}
			child [level 3] { node {eco shower\\head}}
			child [level 3] { node {bathtub\\size}}
			child [level 3] { node {showering\\frequency}}
			child [level 3] { node {electrical\\devices}}
			child [level 3] { node {bathing\\frequency}}
			child [level 3] { node {water\\temperature}}
		}
		child [level 2] { node {cooking}
			child [level 3, rotate=70, level distance=7em] { node {number of\\inhabitants}}
			child [level 3, rotate=70, level distance=8em] { node {gender}}
			child [level 3, rotate=70, level distance=7em] { node {geolocation}}
			child [level 3, rotate=70, level distance=7em] { node {culture}}
		}
	}
;
\end{tikzpicture}
}
\caption{A taxonomy of features relevant for energy application data model design.}
\label{fig:ApplicableFeatures}
\end{figure*}
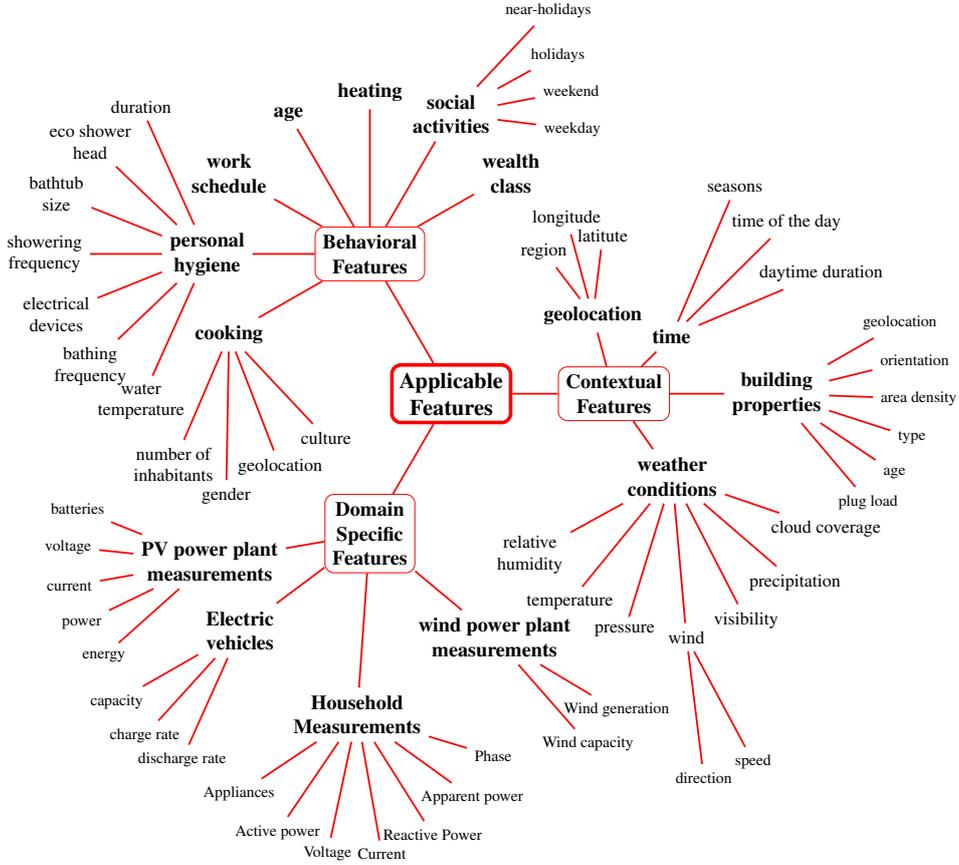

In this section, we propose a taxonomy that identifies and structures various types of data related to energy applications in view of data model development for ML applications in energy. The proposed taxonomy is depicted in Figure~\ref{fig:ApplicableFeatures} and distinguishes three large categories: domain specific, contextual and behavioural. To achieve inter-database or inter-feature store interoperability, the proposed model can be encoded in semantically inter-operable formats using already available vocabularies and ontologies such as \ac{saref}, \ac{sargon} and Schema.org for the domain specific features, \ac{owl}, \ac{foaf} and other semantic structures available to create linked open datasets\footnote{Linked Open Data Vocabularies, https://lov.linkeddata.es/dataset/lov/vocabs} ontologies for behavioral and contextual features.  

\begin{table*}[htbp]
\footnotesize
\caption{Datasets and studies suitable for the categories in the feature taxonomy.}
\label{tab:datasets}
\centering
\begin{tabular}{|lllll|}
    \toprule
    \multicolumn{5}{|c|}{Domain specific features}
    \\

    \thead{PVs} & \thead{EVs} & \thead{Wind} & \multicolumn{2}{c|}{Household}
    \\\midrule

    \makecell[tl]{
        UKPN~\cite{UKPN}\\
        Kannal~\cite{KannalIN}
    }
    &
    \makecell[tl]{
        V2G~\cite{soares2013v2g}\\
        emobpy~\cite{morales2021emobpy}
    }
    &
    \makecell[tl]{
        Sandoval~\cite{SandovalGER}\\
        Lafaz~\cite{LafazGER}
    }
    &
    \makecell[tl]{
        BLOND-50~\cite{BLOND}\\
        FIRED~\cite{FIRED}\\
        UK-DALE~\cite{UKDALE}\\
        REFIT~\cite{REFIT}\\
        ECO~\cite{ECO}\\
        REDD~\cite{REDD}
    }
    &
    \makecell[tl]{
        iAWE~\cite{iAWE}\\
        COMBED~\cite{COMBED}\\ % !
        HUE~\cite{makonin2018hue}\\
        HVAC~\cite{grigore2019hvac}\\
        Synthetic~\cite{li2021synthetic}\\
        UCI Household~\cite{UCIML2012household}
    }

\\\bottomrule
\end{tabular}

\begin{tabular}{|lll||l|}
    \toprule
    \multicolumn{3}{|c||}{Contextual features}
    & Behavioral features
    \\
    & \multicolumn{2}{c||}{Weather} & Building
    \\\midrule

    \makecell[tl]{
        UKPN~\cite{UKPN}\\
        Kannal~\cite{KannalIN}\\
        Sandoval~\cite{SandovalGER}\\
        Lafaz~\cite{LafazGER}
    }
    &
    \makecell[tl]{
        ECN~\cite{UK-ECN}\\
        MIDAS~\cite{MIDAS}\\
        Keller\cite{KellerGER}\\
        Kukreja~\cite{KukrejaIN}
    }
    &
    \makecell[tl]{
        HUE~\cite{makonin2018hue}\\
        REFIT~\cite{REFIT}
    }
    &
    \makecell[lt]{
        Social~\cite{Spichakova2019FeatureEngineering}\\
        Wealth~\cite{Adika2012NorthAmericanPowerSymposium(NAPS)}\\
        Gender~\cite{parkinson2021overcooling}
    }

\\\bottomrule
\end{tabular}
\end{table*}
\subsection{Domain specific}
Domain-specific features are measurements of energy consumption and production collected by \acp{ied} installed at the various points of the energy grid. Additionally, information associated with energy-related appliances, such as the type of heating (\ie, heat pump, gas furnace, electric fireplace), can be presented in meta-data. In Figure~\ref{fig:ApplicableFeatures} we identify \ac{pv} power plant generation, \acp{ev}, wind power generation, and household consumption. Power plants data include battery/super-capacitor capacity, voltage, current, power, and energy measurements and can be found in at least two publicly available datasets as listed in Table~\ref{tab:datasets}. These datasets may also include metadata such as plant id, source key, geographical location, and measurements that fall under the contextual features group, such as air temperature. Wind power generation datasets contain the generated power and sometimes electrical capacity.

%With a recent spike in the popularity of \acp{ev}, power grids need to be designed/improved accordingly because of their significant energy capacity and power draw. 

In the third column of Table~\ref{tab:datasets}, we listed good quality publicly accessible datasets related to \acp{ev}. The datasets consider, for instance, battery capacity, charge rate, and discharge rate. For household measurements, as per the fourth column of Table~\ref{tab:datasets}, consumption may be measured by a single or several smart meters, thus being aggregated or per (set of) appliances. Depending on the dataset, they contain active, reactive, and apparent power, current, phase, and voltage. 
%In some cases, meta-data about the geolocation, orientation, or size of the house may also be present. There are a number of good quality publicly accessible datasets, as can be seen from the fourth column of Table~\ref{tab:datasets}.

\subsection{Contextual features}

We refer to contextual features as measured data that is not directly collected by measuring device(s), however such data~\cite{Sinimaa2021FeatureEngineering} may be critical in developing better energy consumption or production estimates as also shown in the analysis in Section~\ref{sec:rw}. In Figure~\ref{fig:ApplicableFeatures} we identify 1) weather data such as wind speed/direction, temperature, relative humidity, pressure, cloud coverage and visibility, 2) building properties such as type of insulation, year of building, type of property, orientation, area density and 3) time related features such as part of the day and daytime duration. 

Weather datasets, such as the ones listed in the fifth column of Table~\ref{tab:datasets}, usually provide numbers related to temperature and precipitation. Some also provide more climate elements like fog and hail, wind speed, humidity, pressure and sunshine/solar data. They also all provide the geographical location and the time period of measurement.

\subsection{Behavioral features}

Behavioral features refer to aspects related to people's behavior such as work schedule, age, wealth class and hygiene habits as per Figure~\ref{fig:ApplicableFeatures}. Several studies, such as~\cite{Spichakova2019FeatureEngineering, Adika2012NorthAmericanPowerSymposium(NAPS), parkinson2021overcooling} show that electricity consumption is influenced by the behavior of the inhabitants. These are, for instance, personal hygiene, person's age, person's origins, work habits, cooking habits, social activities~\cite{Spichakova2019FeatureEngineering}, and wealth class~\cite{Adika2012NorthAmericanPowerSymposium(NAPS)}. People of different ages have diverse electricity consumption patterns because of the difference in sleep habits and lifestyles in general. Social activities such as holidays, near-holidays, weekdays, and weekends also make a difference in the electricity consumption in homes, offices, and other buildings. Wealth class also influences electricity consumption because it impacts the lifestyle. If we are talking about a home, personal hygiene also has a major impact. Showering or bathing can take a considerable amount of warm water, depending on the showering frequency, shower duration, and how much water the shower head pours each minute. Besides that, we have to consider electrical devices such as a fan, hairdryer, or infrared heater.

Gender of the inhabitants also plays a role as men and women have different preferences when it comes to air temperature, water temperature~\cite{parkinson2021overcooling} and daily routines. While such data is useful for estimating and optimizing energy consumption, it raises ethical and privacy concerns. To some extent, such data can be collected through questionnaires, studies, or simulations to describe the behavioral specifics of users, groups, or communities.

\section{Methodology and Experiment Details}
\label{sec:methodology}

% TODO: Improve the following text.
To validate the proposed taxonomy in Section~\ref{sec:taxonomy} to facilitate the development of comprehensive feature sets, this study empirically evaluates the importance of an extensive list of features and their impact on a household electricity forecasting application using machine learning models. %In the following subsections, we present dataset details, train-test split details, data preparation, algorithms, model training and feature importance analysis.

\subsection{Datasets}
\label{sub:datasets}

For our experimental evaluation, we selected three openly available and widely used datasets, namely, UCI Household~\cite{UCIML2012household}, HUE~\cite{makonin2018hue}, and REFIT~\cite{REFIT} dataset, for energy forecasting. All power-related time series are converted to energy with an hourly sampling rate, which is the lowest common denominator, for meaningful comparison between datasets later. The target values is energy use forecast an hour ahead. % The goal is to forecast energy 1h in advance.

The HUE dataset contains records of 28 households with three or more years of collected data. The raw data (see Table~\ref{tab:dataset-features}, second row) includes domain-specific features such as hourly energy consumption, \ac{ev} battery capacity, and weather conditions, including temperature, pressure, humidity, and cloudiness. Additionally, supporting metadata contains contextual features such as household orientation, type, and heating/cooling infrastructure. Due to lack of sub-meter data, this dataset contains the least amount of behavioural features of the three datasets utilised in the experiment.

The UCI Household dataset includes multivariate time-series data collected over 47 months from one household. The raw dataset, as shown in Table~\ref{tab:dataset-features} in the first row, comprises domain-specific features such as absolute timestamp, active and reactive power, voltage, and electrical current from the main meter, as well as three additional sub-meters. Furthermore, the UCI Household metadata includes contextual features such as approximate geolocation. Through the sub-meter data, here additional behavioural features could be extracted, such as kitchen activity.

The REFIT dataset contains records of 20 households with a sampling rate of 8 seconds, recorded between 2013 and 2015. The raw dataset (see Table~\ref{tab:dataset-features}, third row) contains domain-specific features such as timestamp and power draw of nine individual appliances and total household power consumption. Additionally, supporting metadata provides contextual features, such as household id, type, construction year, size, number of residents, and total appliances owned. Due to more rich sub-meter data and higher number of connected appliances, we could further enhance the behavioural features set compared to the other two datasets, such as such as personal grooming, cooking, cleaning, entertainment, bedroom activities, and work-at-home activities.
\begin{table*}[htbp]
\caption{Available feature in dataset used for evaluation.}
\label{tab:dataset-features}
\centering
%\ra{1.3}
\footnotesize
\begin{tabularx}{\hsize}{@{}l|X|X|X@{}}
    \multirow{2}{*}{\thead{Dataset}} & \multicolumn{3}{c}{Features} \\\cline{2-4}
    & \thead{Domain specific}
    & \thead{Contextual}
    & \thead{Behavioral}
    \\
    \midrule\midrule

    \makecell{HUE\,\cite{makonin2018hue}}
    &
    \textbf{Measurements:} energy consumption, rolling average and standard deviation over time window, consumption \{23, 24\} hours ago, consumption a week ago \newline
    \textbf{Electrical vehicle:} EV battery size,
    &
    \textbf{Geolocation:} latitude, longitude, country, region, household identifier\newline
    \textbf{Time:} timestamp, part of \{day, week, year\}, timestamp, timezone, daylight saving time (DST)\newline
    \textbf{Weather conditions:} humidity, temperature, air pressure, clouds, solar \{altitude, azimuth, radiation\}\newline
    \textbf{Building properties:} building type, building orientation, \# rental units, air condition system, air gas furnace, heat pump, gas fireplace, electric fireplace, in floor heating, portable air conditioner, cast iron radiators, geothermal heating
    &
    \textbf{social activities:} weekday, holidays, weekend.\newline
    \textbf{cooking:} breakfast (6-9), lunch (11-15), dinner (18-21).\newline
    \textbf{work schedule:} yesterday consumption ratio, yesterday median consumption, house activity metadata, work schedule (9-17), free time schedule (17-22), sleep schedule (22-7)
    \\\midrule
    \makecell{UCI\newline\\Household\\\cite{UCIML2012household}}
    &
    \textbf{Measurements:} total active power, total reactive power, voltage, electric current, 3 submeters, rolling average and standard deviation over observation window, consumption \{23, 24\} hours ago, consumption a week ago
    &
    \textbf{Geolocation:} latitude, longitude, region, country \newline \textbf{Time:} part of \{day, week, year\}, timestamp, timezone, daylight saving time (DST)\newline
    \textbf{Weather condition:} solar \{altitude, azimuth, radiation\}
    &
    \textbf{cooking:} kitchen activity (submeter \#1), breakfast (6-9), lunch (11-15), dinner (18-21).\newline
    \textbf{social activities:} weekday, holidays, weekend.\newline
    \textbf{work schedule:} yesterday consumption ratio, yesterday median consumption, house activity metadata, work schedule (9-17), free time schedule (17-22), sleep schedule (22-7)
    
    \\\midrule
    
    \makecell{REFIT\,\cite{REFIT}}
    &
    \textbf{Measurements:} total active power, 9 selected appliances, rolling average and standard deviation over time window, consumption \{23, 24\} hours ago, consumption a week ago\newline
    &
    \textbf{Geolocation:} latitude, longitude, country, region, household identifier\newline
    \textbf{Time:} timestamp, part of \{day, week, year\}, timestamp, timezone, daylight saving time (DST)\newline
    \textbf{Weather conditions:} solar \{altitude, azimuth, radiation\} \newline
    \textbf{Building properties:} \# residents, building type, household size, \# appliances owned, contruction year
    & 
    \textbf{cooking:} kitchen activity (based on kitchen appliances), breakfast (6-9), lunch (11-15), dinner (18-21)\newline
    \textbf{social activities:} weekday, holidays, weekend, personal grooming, cleaning, entertainment, work at home, heating, bedroom activity.\newline
    \textbf{work schedule:} yesterday consumption ratio, yesterday median consumption, house activity metadata, work schedule (9-17), free time schedule (17-22), sleep schedule (22-7)
    \\
    \bottomrule
\end{tabularx}
\end{table*}

% To further enhance the available features of the datasets considered, we added the following supplementary features:

% \begin{itemize}
%     \item Rolling averages and standard deviations of the previous three hours for all domain-specific time series features.
%     \item Contextual features that depend on location and absolute time, such as time of day, day of the week and year, theoretical clear-sky solar radiation, solar altitude, and solar azimuth.
%     \item Behavioral features related to kitchen or house activities, social activities such as weekdays, holidays, and weekends, as well as (relative) energy consumption 24 and 23 hours ago, consumption exactly one week ago, and yesterday's median consumption.
% \end{itemize}

\subsection{Data preparation, algorithms, model training}

%To show the relevance of domain-specific, contextual, and behavioral features, we empirically evaluate the contribution of each feature group to the forecast model.

For evaluation, we prepared the data for model training as follows. First, data is split for each household into train and test subsets in an 80:20 ratio. This way, we always forecast the last 20\% of the time series. We do not shuffle the data to prevent any potential data leakage. Second, tabular algorithms (\ac{lr}, \ac{xgb}, \ac{mlp}) train on combined train sets from all households, while multivariate algorithms (Prophet, N-BEATS) are trained for each household separately due to algorithms' constraints. Third, we build a model and evaluate performance with \ac{mpe} defined as

\begin{equation}
    \textrm{MPE} = \frac{1}{N} \sum^{N}_{i=0} \frac{|A_i - F_i|}{\max (|A_i|, |F_i|, \varepsilon)} \cdot 100\%
\end{equation}

and \ac{mse} defined as

\begin{equation}
    \textrm{MSE} = \frac{1}{N}\sum_{i=0}^{N}(A_i - F_i)^2
\end{equation}

metrics, where $A_i$ represents the actual value and $F_i$ denotes forecast value for $i$-th sample. The \ac{mpe} metric is particularly sensitive to cases where both $A_i$ and $F_i$ approach zero. To mitigate the risk of division by zero and consequent instability in the \ac{mpe} calculation, we introduce a small positive constant, $\varepsilon$, into the denominator. We then select the maximum value among  $A_i$, $F_i$, and $\varepsilon$ for the denominator, ensuring that the metric remains robust and avoids unbounded values even when actual and forecast values are extremely low.

For the evaluation, we selected five machine learning algorithms. \Acf{lr}, \acf{xgb} regressor, \acf{mlp} as tabular data algorithms, Prophet and N-BEATS as multivariate time-series algorithms. All algorithms use default settings\footnote{scikit-learn v1.1, XGBoost v1.7.5, DARTS v0.23.1, and Prophet v1.1.2} of their respective implementations with the following exceptions. For the \ac{mlp} regressor, we increased the number of iterations from 200 to 500, and the learning rate is set to \textit{adaptive}. For N-BEATS, we use the same configuration as in~\cite{oreshkin2021grid}, where authors used three blocks with three layers, layer width is 512 neurons, an input sequence of length 32, an output sequence of length one, and mean absolute percentage error as loss metric.

%% TODO: Some more text about algorithms ...

\subsection{Feature Importance Analysis}
\label{sub:methodology:shap}

Following the proposed taxonomy and discussion in Section~\ref{sec:taxonomy}, we analyze and evaluate the relevance of the features. Firstly, we empirically assess feature groups' contribution to the forecast model. The features are divided into three groups, as defined in Figure~\ref{fig:ApplicableFeatures} at the first level of distinction: domain-specific, contextual, and behavioral features. Secondly, we examine the contributions of individual features using the tree-based \ac{shap} metric. Finally, we present and compare the cumulative \ac{shap} scores of the three feature groups. The analysis is per dataset, with and without sub-meters, with common conclusions.

We used \ac{shap}, a unified framework, to interpret the prediction of any machine learning model. It connects optimal credit allocation with local explanations using the traditional Shapley values from game theory and their related extensions \cite{lundberg2017shap, lundberg2020local2global}. The framework assigns each feature an importance value for a particular prediction. Usually, an average of these values for all (or a subset of) samples presents overall feature importance. Unlike other methods, such as permutation importance or impurity-based importance, \ac{shap} can better recognize the feature importance of dependent or highly correlated features. Furthermore, \cite{baudeu2023shap} shows that \ac{shap} is less biased towards features with high cardinality if the data has a reasonable signal-to-noise ratio.

\section{Results}
\label{sec:evaluation}

In this section, we present and analyze the results of the experiments defined in Section~\ref{sec:methodology}. Section~\ref{sub:evaluation:models} evaluates the contributions to the forecasting model of three feature groups, using five machine learning algorithms on three different datasets. %Beside seven combinations of feature groups, we also evaluate two special cases: using only raw data, and excluding sub-metered data. 
Sections~\ref{sub:evaluation:models} and~\ref{sub:evaluation:groups} analyze feature group and individual feature contribution to the forecast accuracy with the \ac{shap} metric.

\subsection{Contribution of the Applicable Feature Groups}
\label{sub:evaluation:models}

% Code: Re forecasting, it is this one: https://github.com/sensorlab/feature-store-energy

\begin{table*}[htbp]
\caption{Energy consumption estimation 1h ahead; HUE dataset. Estimating value in kWh.}
\label{tab:hue:prediction}
\ra{1.2}
\centering
\footnotesize
\begin{tabular}{r | ccc | rr | rr | rr | rr | rr}
    %\toprule
    & \multicolumn{3}{c}{Feature groups (see Table~\ref{tab:dataset-features})}
    & \multicolumn{2}{c}{\ac{xgb} regr.} 
    & \multicolumn{2}{c}{Linear regr.}
    & \multicolumn{2}{c}{MLP}
    & \multicolumn{2}{c}{Prophet} 
    & \multicolumn{2}{c}{N-BEATS} 
    \\
    & \thead{Domain} & \thead{Contextual} & \thead{Behavioral}
    & \thead{MSE} & \thead{MPE}
    & \thead{MSE} & \thead{MPE}
    & \thead{MSE} & \thead{MPE}
    & \thead{MSE} & \thead{MPE}
    & \thead{MSE} & \thead{MPE}
    \\
    &&&& \thead{kWh} & \thead{\%} & \thead{kWh} & \thead{\%} & \thead{kWh} & \thead{\%} & \thead{kWh} & \thead{\%} & \thead{kWh} & \thead{\%}
    \\\midrule

    {\color{gray}1} &
    \checkmark* & & 
    & 0.417 & 28.406  % 2023-04-20 -- XGB
    & 0.418 & 30.839  % 2023-04-20 -- Linear
    & 0.410 & 28.655  % 2023-04-20 -- MLP
    & 0.424 & 31.890  % 2023-04-20 -- Prophet
    & 0.815 & 71.242  % 2023-04-20 -- N-BEATS
    \\\midrule
    
    {\color{gray}2} &
    \checkmark & &
    & 0.335 & 25.276  % 2023-04-20 -- XGB
    & 0.352 & 27.031  % 2023-04-20 -- Linear
    & 0.324 & 25.174  % 2023-04-20 -- MLP
    & 0.325 & 28.077  % 2023-04-20 -- Prophet
    & 0.884 & 78.026  % 2023-04-20 -- N-BEATS
    \\

    {\color{gray}3} &
    & \checkmark &
    & 0.519 & 33.305  % 2023-04-20 -- XGB
    & 0.651 & 41.731  % 2023-04-20 -- Linear
    & 0.528 & 36.446  % 2023-04-20 -- MLP
    & 0.669 & 38.755  % 2023-04-20 -- Prophet
    & 0.890 & 78.441  % 2023-04-20 -- N-BEATS
    \\

    {\color{gray}4} &
    && \checkmark
    & 0.439 & 29.210  % 2023-04-20 -- XGB
    & 0.529 & 37.069  % 2023-04-20 -- Linear
    & 0.436 & 28.929  % 2023-04-20 -- MLP
    & 0.545 & 35.596  % 2023-04-20 -- Prophet
    & 0.884 & 77.514  % 2023-04-20 -- N-BEATS
    \\\midrule

    {\color{gray}5} & \checkmark & \checkmark & 
        & \textbf{0.316} & \textbf{24.731}  % 2023-04-20 -- XGB
    & 0.348 & 26.952  % 2023-04-20 -- Linear
    & 0.312 & 27.476  % 2023-04-20 -- MLP
    & 0.322 & 28.120  % 2023-04-20 -- Prophet
    & 0.894 & 78.483  % 2023-04-20 -- N-BEATS
    \\

    {\color{gray}6} &
    & \checkmark & \checkmark
    & 0.402 & 28.418  % 2023-04-20 -- XGB
    & 0.511 & 38.555  % 2023-04-20 -- Linear
    & 0.430 & 31.756  % 2023-04-20 -- MLP
    & 0.477 & 34.697  % 2023-04-20 -- Prophet
    & 0.881 & 77.591  % 2023-04-20 -- N-BEATS
    \\

    {\color{gray}7} &
    \checkmark && \checkmark
    & 0.325 & 24.986  % 2023-04-20 -- XGB
    & 0.345 & 26.713  % 2023-04-20 -- Linear
    & 0.315 & 25.178  % 2023-04-20 -- MLP
    & 0.320 & 27.585  % 2023-04-20 -- Prophet
    & 0.871 & 76.663  % 2023-04-20 -- N-BEATS
    \\\midrule

    {\color{gray}8} &
    \checkmark & \checkmark & \checkmark
    & \textbf{0.316} & \textbf{24.675}  % 2023-04-20 -- XGB
    & 0.344 & 27.285  % 2023-04-20 -- Linear
    & 0.519 & 62.659  % 2023-04-20 -- MLP
    & 0.320 & 28.034  % 2023-04-20 -- Prophet
    & 0.887 & 77.877  % 2023-04-20 -- N-BEATS
    \\
    
\end{tabular}
\end{table*}

\begin{table*}[htbp]
\caption{Energy consumption estimation 1h ahead; UCI Household dataset. Estimating value in kWh.}
\label{tab:uciml:prediction}
\ra{1.2}
\centering
\footnotesize
\begin{tabular}{r | ccc | rr | rr | rr | rr | rr}
    %\toprule
    & \multicolumn{3}{c}{Feature groups (see Table~\ref{tab:dataset-features})}
    & \multicolumn{2}{c}{\ac{xgb} regr.} 
    & \multicolumn{2}{c}{Linear regr.}
    & \multicolumn{2}{c}{MLP}
    & \multicolumn{2}{c}{Prophet}
    & \multicolumn{2}{c}{N-BEATS} 
    \\
    & \thead{Domain} & \thead{Contextual} & \thead{Behavioral}
    & \thead{MSE} & \thead{MPE}
    & \thead{MSE} & \thead{MPE}
    & \thead{MSE} & \thead{MPE}
    & \thead{MSE} & \thead{MPE}
    & \thead{MSE} & \thead{MPE}
    \\
    &&&& \thead{kWh} & \thead{\%} & \thead{kWh} & \thead{\%} & \thead{kWh} & \thead{\%} & \thead{kWh} & \thead{\%} & \thead{kWh} & \thead{\%}
    \\\midrule
    
    {\color{gray}1} &
    \checkmark* & &
    & 0.292 & 31.052  % 2023-04-19 -- XGB
    & 0.297 & 33.410  % 2023-04-19 -- Linear
    & 0.302 & 34.154  % 2023-04-19 -- MLP
    & 0.261 & 31.438  % 2023-04-19 -- Prophet
    & 1.115 & 50.705  % 2023-04-19 -- N-BEATS
    \\\midrule

    {\color{gray}2} &
    \checkmark & &
    & 0.233 & 27.078  % 2023-04-19 -- XGB
    & 0.247 & 29.832  % 2023-04-19 -- Linear
    & 0.657 & 77.520  % 2023-04-19 -- MLP
    & 0.238 & 29.778  % 2023-04-19 -- Prophet
    & 1.066 & 49.945  % 2023-04-19 -- N-BEATS
    \\

    {\color{gray}3} & 
    & \checkmark &
    & 0.407 & 36.452  % 2023-04-19 -- XGB
    & 0.560 & 45.991  % 2023-04-19 -- Linear
    & 0.655 & 51.281  % 2023-04-19 -- MLP
    & 0.424 & 42.155  % 2023-04-19 -- Prophet
    & 1.057 & 57.859  % 2023-04-19 -- N-BEATS
    \\

    {\color{gray}4} & 
    && \checkmark
    & 0.422 & 36.283  % 2023-04-19 -- XGB
    & 0.393 & 37.319  % 2023-04-19 -- Linear
    & 0.385 & 36.544  % 2023-04-19 -- MLP
    & 0.376 & 38.575  % 2023-04-19 -- Prophet
    & 1.280 & 75.109  % 2023-04-19 -- N-BEATS
    \\\midrule

    {\color{gray}5} & 
    \checkmark & \checkmark & 
    & \textbf{0.222} & \textbf{26.641}  % 2023-04-19 -- XGB
    & 0.246 & 29.966  % 2023-04-19 -- Linear
    & 0.353 & 36.439  % 2023-04-19 -- MLP
    & 0.238 & 29.705  % 2023-04-19 -- Prophet
    & 1.044 & 48.751  % 2023-04-19 -- N-BEATS
    \\

    {\color{gray}6} & 
    & \checkmark & \checkmark
    & 0.374 & 34.352  % 2023-04-19 -- XGB
    & 0.383 & 37.081  % 2023-04-19 -- Linear
    & 0.415 & 38.051  % 2023-04-19 -- MLP
    & 0.373 & 37.768  % 2023-04-19 -- Prophet
    & 1.570 & 100.467  % 2023-04-19 -- N-BEATS
    \\

    {\color{gray}7} & 
    \checkmark && \checkmark
    & \textbf{0.222} & \textbf{26.796}  % 2023-04-19 -- XGB
    & 0.237 & 29.482  % 2023-04-19 -- Linear
    & 0.299 & 37.520  % 2023-04-19 -- MLP
    & 0.235 & 29.798  % 2023-04-19 -- Prophet
    & 1.495 & 89.918  % 2023-04-19 -- N-BEATS
    \\\midrule

    {\color{gray}8} & 
    \checkmark & \checkmark & \checkmark
    & \textbf{0.223} & \textbf{26.346}  % 2023-04-19 -- XGB
    & 0.236 & 29.576  % 2023-04-19 -- Linear
    & 0.511 & 63.373  % 2023-04-19 -- MLP
    & 0.235 & 29.732  % 2023-04-19 -- Prophet
    & 1.060 & 49.710  % 2023-04-19 -- N-BEATS
    \\

    \midrule
    {\color{gray}9} & 
    \multicolumn{3}{c|}{\checkmark\checkmark\checkmark (without submeter data)}
    & 0.252 & 27.846  % 2023-04-19 -- XGB
    & 0.260 & 29.863  % 2023-04-19 -- Linear
    & 0.353 & 37.295  % 2023-04-19 -- MLP
    & 0.252 & 30.002  % 2023-04-19 -- Prophet
    & 2.151 & 138.139  % 2023-04-19 -- N-BEATS
    \\
    
\end{tabular}
\end{table*}

\begin{table*}[htbp]
\caption{Energy consumption estimation 1h ahead; REFIT dataset. Estimating value in kWh.}
\label{tab:refit:prediction}
\ra{1.2}
\centering
\footnotesize
\begin{tabular}{r | ccc | rr | rr | rr | rr | rr}
    & \multicolumn{3}{c}{Feature groups (see Table~\ref{tab:dataset-features})}
    & \multicolumn{2}{c}{\ac{xgb} regr.} 
    & \multicolumn{2}{c}{Linear regr.}
    & \multicolumn{2}{c}{MLP}
    & \multicolumn{2}{c}{Prophet} 
    & \multicolumn{2}{c}{N-BEATS} 
    \\
    & \thead{Domain} & \thead{Contextual} & \thead{Behavioral}
    & \thead{MSE} & \thead{MPE}
    & \thead{MSE} & \thead{MPE}
    & \thead{MSE} & \thead{MPE}
    & \thead{MSE} & \thead{MPE}
    & \thead{MSE} & \thead{MPE}
    \\
    &&&& \thead{kWh} & \thead{\%} & \thead{kWh} & \thead{\%} & \thead{kWh} & \thead{\%} & \thead{kWh} & \thead{\%} & \thead{kWh} & \thead{\%}
    \\\midrule

    {\color{gray}1} &
    \checkmark* & & 
    & 0.148 & 29.691  % 2023-04-20 -- XGB
    & 0.212 & 36.947  % 2023-04-20 -- linear
    & 0.163 & 34.408  % 2023-04-20 -- MLP
    & 0.197 & 36.073  % 2023-04-20 -- Prophet
    & 1.125 & 45.647  % 2023-04-20 -- N-BEATS
    \\\midrule
    
    {\color{gray}2} &
    \checkmark & &
    & 0.132 & 25.934  % 2023-04-20 -- XGB
    & 0.145 & 28.866  % 2023-04-20 -- linear
    & 0.132 & 25.530  % 2023-04-20 -- MLP
    & 0.153 & 31.627  % 2023-04-20 -- Prophet
    & 0.299 & 39.128  % 2023-04-20 -- N-BEATS
    \\

    {\color{gray}3} &
    & \checkmark &
    & 0.407 & 35.321  % 2023-04-20 -- XGB
    & 0.402 & 42.020  % 2023-04-20 -- linear
    & 1.894 & 71.698  % 2023-04-20 -- MLP
    & 0.506 & 55.495  % 2023-04-20 -- Prophet
    & 1.153 & 47.100  % 2023-04-20 -- N-BEATS
    \\

    {\color{gray}4} &
    && \checkmark
    & 0.183 & 30.719  % 2023-04-20 -- XGB
    & 0.188 & 33.339  % 2023-04-20 -- linear
    & 0.180 & 31.058  % 2023-04-20 -- MLP
    & 0.211 & 37.283  % 2023-04-20 -- Prophet
    & 0.317 & 40.258  % 2023-04-20 -- N-BEATS
    \\\midrule

    {\color{gray}5} &
    \checkmark & \checkmark & 
    & 0.126 & 25.580  % 2023-04-20 -- XGB
    & 0.145 & 28.977  % 2023-04-20 -- linear
    & 0.471 & 59.179  % 2023-04-20 -- MLP
    & 0.155 & 32.022  % 2023-04-20 -- Prophet
    & 0.300 & 39.349  % 2023-04-20 -- N-BEATS
    \\

    {\color{gray}6} &
    & \checkmark & \checkmark
    & 0.259 & 32.621  % 2023-04-20 -- XGB
    & 0.183 & 33.248  % 2023-04-20 -- linear
    & 0.284 & 60.288  % 2023-04-20 -- MLP
    & 0.201 & 40.987  % 2023-04-20 -- Prophet
    & 0.318 & 39.760  % 2023-04-20 -- N-BEATS
    \\

    {\color{gray}7} &
    \checkmark && \checkmark
    & 0.126 & 25.543  % 2023-04-20 -- XGB
    & 0.144 & 27.671  % 2023-04-20 -- linear
    & 0.170 & 27.254  % 2023-04-20 -- MLP
    & 0.151 & 31.651  % 2023-04-20 -- Prophet
    & 0.298 & 38.998  % 2023-04-20 -- N-BEATS
    \\\midrule

    {\color{gray}8} &
    \checkmark & \checkmark & \checkmark
    & \textbf{0.123} & \textbf{25.389}  % 2023-04-20 -- XGB
    & 0.144 & 28.401  % 2023-04-20 -- linear
    & 0.290 & 50.648  % 2023-04-20 -- MLP
    & 0.152 & 32.589  % 2023-04-20 -- Prophet
    & 0.300 & 39.682  % 2023-04-20 -- N-BEATS
    \\
    \midrule
    {\color{gray}9} &
    \multicolumn{3}{c|}{\checkmark\checkmark\checkmark (without submeter data)}
    & 0.133 & 25.726  % 2023-04-20 -- XGB
    & 0.143 & 28.625  % 2023-04-20 -- linear
    & 1.921 & 74.177  % 2023-04-20 -- MLP
    & 0.143 & 32.389  % 2023-04-20 -- Prophet
    & 0.299 & 38.781  % 2023-04-20 -- N-BEATS
    \\
\end{tabular}
\end{table*}

In this subsection, we focus on the contributions of three feature groups to the final forecasting model for three individual datasets. In addition to seven combinations of feature groups, we also evaluate two special cases: using only raw data and excluding sub-metered data. The results are presented in Table~\ref{tab:uciml:prediction} for the UCI Household dataset, in Table~\ref{tab:hue:prediction} for the HUE dataset and in Table~\ref{tab:refit:prediction} for the REFIT dataset.

Tables~\ref{tab:hue:prediction},~\ref{tab:uciml:prediction}, and~\ref{tab:refit:prediction} are structured as follows. Columns 1-3 indicate the combination of feature groups used for the forecasting. The fourth and fifth columns contain the results for the linear regression, the sixth and seventh columns for \ac{xgb} Regressor, the eighth and ninth for \ac{mlp}, the tenth and eleventh for Prophet, and finally the twelfth and thirteenth columns contain the results for N-BEATS. For each algorithm, we measure the \acf{mse} in kilowatt hours and the \acf{mpe} in percent. The first row of data has an asterisk by the domain-specific features meaning that only raw data directly available from the dataset was used. Lines 2-4 report on individual groups of features. Lines 5-7 report on pairs of features, row 8 uses all available features, while row 9 excludes features obtained via submeters (applies to UCI and REFIT).

Table~\ref{tab:hue:prediction} details the results for the HUE dataset that contains the least amount of behavioural features out of the three datasets. It can be observed that the \ac{xgb} regressor outperforms the other algorithms for the majority of feature combinations. A comparison of rows 2 to 4 shows that the models that utilising only domain-specific features perform better than the models that only consider contextual and behavioral features. However, a comparison of the results in rows 3 and 4 shows that the models using behavioral features tend to perform better than those using contextual features. As can be further seen in Table~\ref{tab:hue:prediction}, combining domain features with the other two feature types further improves the performance of the models, with the best performance observed when using all three feature types in row 8. If we compare the \ac{xgb} results in row 8 with those in row 1, we see that feature engineering can improve the performance of the model by $\approx$ 3.73 percentage points with respect to the metric \ac{mpe} compared to just utilising raw time series data.

Table~\ref{tab:uciml:prediction} presents the results for the UCI Household dataset, which already includes additional behavioural features compared to the HUE dataset. Also here we notice that the \ac{xgb} regressor consistently demonstrated superior performance across various feature combinations, particularly when domain and contextual features were included, achieving the lowest \ac{mse} of 0.222 and \ac{mpe} of 26.641\% in row 5 and 7. A similar performance of \ac{xgb} can be seen in row 8, when using all three types of features, where it performs slightly worse in terms of \ac{mse} compared to combination of domain features with either behavioural or contextual, but achieves a slightly better \ac{mpe}. This indicates \ac{xgb}’s robustness in capturing complex interactions inherent in the dataset. Prophet also showed commendable performance with raw domain-specific features in row 1, with an \ac{mse} of 0.261 and \ac{mpe} of 31.438\%. Additionally, row 9 contains results with omitted additional behavioural features. Looking at \ac{xgb}, as the best performing model, it can be seen, the model performs with \ac{mse} of 0.252, which is 0.019 worse compared to the model that utilise these additional behavioural features in row 8. This indicates that additional behavioural features have a meaningful impact on the forecasting performance of the model. Similarly to HUE dataset, comparing the \ac{xgb} results in row 8 to row 1, we can see that feature engineering improves the performance of the model by $\approx$4.70 percentage points in terms of \ac{mpe} metric compared to just raw time series data. This $\approx$1 percentage point better then for HUE, which indicates that additional behavioural features further improve the forecasting performance.

Table~\ref{tab:refit:prediction} shows the results for the REFIT dataset, which contained the largest number of behavioral features. \ac{xgb} regressor again showed the best overall performance. Similar observations can be made as in the previous two tables, with the domain features in row 2 having the largest impact on the performance of the models compared to the contextual and behavioral features, with the behavioral features clearly outperforming the contextual ones. Again, the best performance with all features combined is observed in row 8, with the lowest \ac{mse} of 0.123 and \ac{mpe} 25.389\%. It can also be seen that the \ac{xgb} model without additional behavioral features performs 0.01 worse in row 9 of Table~\ref{tab:refit:prediction} compared to the model with all behavioral features. Comparing the results of \ac{xgb} with all types of features in row 8 with the model with only raw time series data in row 1 shows that in terms of \ac{mpe}, model with features outperforms the model with raw time series by $\approx $4.30 percentage points.

The findings from the three datasets in Tables~\ref{tab:hue:prediction},~\ref{tab:uciml:prediction}, and~\ref{tab:refit:prediction}, show that engineering features guided by the proposed taxonomy in Section~\ref{sec:taxonomy}, can significantly improve the forecasting results compared to just utilising raw electric consumption measurement data, which is consistent with findings in~\cite{mostafa2022covid}. Additionally, it shows that additional behavioural features improve the performance of forecast, which is consistent with findings in \cite{Spichakova2019FeatureEngineering}, that electrical consumption is influenced by the behavior of consumers.

The results across the three datasets in Tables~\ref{tab:hue:prediction},~\ref{tab:uciml:prediction}, and~\ref{tab:refit:prediction},  also highlight the varying capabilities of the models in handling different types of feature sets. The results highlight that although different types of features can increase the forecasting performance of models it is equally important to select appropriate models based on the nature of the features and the specific dataset characteristics. For example, Prophet and N-BEATS were designed to mostly work on raw time series data, while the feature set constructed through the taxonomy proposed in Section~\ref{sec:taxonomy} can be considered as tabular data. Due to this, \ac{xgb} regressor consistently outperformed other models, particularly with complex, non-linear feature interactions, as evidenced by its superior performance across most feature combinations and datasets. This underscores its robustness and adaptability in diverse scenarios. Our results additionally support the findings from~\cite{miller2021limitations} where they showed that large ensembles of mainly gradient boosting trees with preprocessing of the training data tend to be the best solutions for forecasting applications.

\subsection{Feature Group Importance Analysis}
\label{sub:evaluation:groups}

\begin{table*}[htbp]
    \caption{SHAP contributions of feature groups.}
    \label{tab:features:contributions}
    \ra{1.3}
    \centering
    \begin{tabular}{l | r || r r r || rrr}
    \multirow{2}{*}{Dataset} & \multicolumn{4}{c||}{number of features} & \multicolumn{3}{c}{cum. rel. SHAP contribution [\%]} \\
    & \thead{Total} & \thead{Domain} & \thead{Contextual} & \thead{Behavioral} & \thead{Domain} & \thead{Contextual} & \thead{Behavioral} \\\midrule

    HUE & 67 & 8 & 44 & 15
    & \textbf{66.5} & 23.0 & 10.5 \\
    
    %HUE (common-only) & 30 & 7 & 9 & 14& \textbf{75.7} & 11.7 & 12.6 \\

    %HUE (top 10\% feat. per category) & 7 & 1 & 4 & 2& 17.4 & 26.8 & 55.8 \\
    
    \midrule
    
    UCI & 88 & 63 & 9 & 16
    & \textbf{83.8} & 9.9 & 6.3 \\

    UCI (no sub-meter) & 39 & 21 & 9 & 9
    & \textbf{73.6} & 20.6 & 5.8 \\

    %UCI (common-only) & 30 & 7 & 9 & 14& \textbf{62.4} & 21.4 & 16.2 \\

    %UCI (less behavioral) & 81 & 63 & 9 & 9 & \textbf{86.6} & 10.5 & 2.9 \\

    %UCI (top 10\% feat. per category) & 9 & 6 & 1 & 2& 70.0 & 6.8 & 23.2 \\

    \midrule
    
    REFIT & 74 & 35 & 18 & 21
    & \textbf{68.2} & 17.0 & 14.8 \\

    REFIT (no sub-meter) & 38 & 7 & 17 & 14
    & \textbf{67.0} & 22.7 & 10.3 \\

    %REFIT (common-only) & 30 & 7 & 9 & 14& \textbf{74.1} & 11.4 & 14.5 \\

    %REFIT (no context from submeters) & 73 & 35 & 17 & 21 & 70.2 & 15.0 & 14.8 \\

    %REFIT (no behavioral from submeters) & 61 & 35 & 18 & 8 & 77.7 & 18.5 & 3.8 \\

    %REFIT (top 10\% feat. per category) & 6 & 3 & 1 & 2& 97.7 & 1.3 & 1.0 \\

    \bottomrule
    \end{tabular}
\end{table*}

% Total sum: D=43,467.015	 C=5,625.046	 B=3,215.461
%Normalized: D=8,309.898	 C=1,075.380	 B=614.723
%83.1 & 10.8 & 6.1 \\

\begin{comment}
\begin{table*}[htbp]
    \caption{SHAP contributions of feature groups.}
    \label{tab:features:contributions}
    \ra{1.3}
    \centering
    \begin{tabular}{l | c c c }
    \multirow{2}{*}{Dataset} & \multicolumn{3}{c}{Cumulative SHAP contribution [\%]} \\
    & \thead{Domain} & \thead{Contextual} & \thead{Behavioral} \\\midrule

    HUE 
    %& 0.5904 & 0.2132 & 0.0984 \\ % These are the original SHAP values
    & \textbf{65.5} & 23.6 & 10.9 \\
    \midrule
    UCI Household 
    %& 1.3012 & 0.1684 & 0.0963 \\ % These are the original SHAP values
    & \textbf{83.} &. & 6. \\

    UCI Household (no sub-meter) 
    %& 0.905 & 0.253 & 0.071 \\ % These are the original SHAP values
    & \textbf{73.6} & 20.6 & 5.8 \\

    \midrule
    
    REFIT 
    %& 0.4807 & 0.1247 & 0.0921 \\ % These are the original SHAP values
    & \textbf{68.9} & 17.9 & 13.2 \\

    REFIT (no sub-meter) 
    %& 0.3568 & 0.1488 & 0.0361 \\ % These are the original SHAP values
    & \textbf{65.8} & 27.5 & 6.7 \\

    \bottomrule
    \end{tabular}
\end{table*}
\end{comment}

%In this section, we follow the methodology described from Section~\ref{sec:methodology} and estimate \ac{shap} contribution values for all three datasets. 

We analyze per feature group contributions and present results in Table~\ref{tab:features:contributions} with the approach from Section~\ref{sec:methodology}. The table summarizes the  number of features in datasets and the \ac{shap} contribution of each feature group. The first column lists all three evaluated datasets. In addition, UCI Household and REFIT dataset are evaluated with and without the behavioural features extracted from sub-meter data. The second colomn lists the total number of features available for training the model after engineering guided by the taxonomy and detailed in Table \ref{tab:dataset-features} while columns 3-5 detail the number of features per group. Finally, columns 6-8 list the  percent of \ac{shap} contribution of individual group.

The results show the domain-specific features being highest contributors according to \ac{shap} score, where they contribute $65.5\%$ on HUE, $83.1\%$ on UCI Household, and $68.9\%$ on REFIT dataset, which is consistent with the results in Tables~\ref{tab:hue:prediction},~\ref{tab:uciml:prediction}, and~\ref{tab:refit:prediction}, where utilising only of the three feature types, the domain features resulted in the lowest error rate. 

Contextual data contributed  $23.6\%$, $10.8\%$, and $17.9\%$ to HUE, UCI Household, and REFIT, respectively. Finally, behavioral features contributed  $10.9\%$, $6.1\%$, and $13.2\%$ to the datasets. The results in Table~\ref{tab:features:contributions} show that, on average, contextual data contributes more to the prediction than  behavioural. This is contradictory to the results in Tables~\ref{tab:hue:prediction},~\ref{tab:uciml:prediction}, and~\ref{tab:refit:prediction}, where if only using contextual data, the \ac{xgb} performs with higher error, then if only using behavioural features. However, this indicates that contextual features provide better support to the domain features for the forecasting task. 

If we look at the feature importance for the UCI and REFIT datasets without the use of sub-meter data in rows three and five of Table~\ref{tab:features:contributions}, we can see that contextual features are gaining importance. Contextual features are mostly provided as metadata such as geolocation or building properties and are therefore independent of the sub-meter data. In contrast, behavioral features are strongly influenced by the sub-meter measurements and their importance decreases when these are removed. This suggests that sub-meter data plays an important role in determining behavioral patterns within a household. UCI has a smaller number of sub-meter data, therefore fewer behavioral patterns were extracted from the households and their removal had less impact on the importance than in REFIT. 

The feature importance results show that more sub-metering data can help extract better behavioral patterns from households, which can have a high impact on the forecasting accuracy of future models.

\section{Conclusions}
\label{sec:conclusions}

In this paper, we proposed a taxonomy developed to empower designing data models for emerging ML-based energy applications. Unlike prior works that focused on structuring energy data in various semantic formats for interoperability purposes, this work aims at data model development and subsequent feature engineering. The three main categories of features identified in the taxonomy are: behavioral, contextual, and domain-specific.

Using consumption forecasting examples on three diverse datasets, we demonstrate the importance of behavioral and contextual features. By adding contextual or behavioral features to feature feature-engineered domain-specific features, we can achieve up to 5\% \ac{mse} and up to 3\% \ac{mpe} improvement. When all three groups of features are combined, we see up to 6\% \acf{mse} and up to 23\% \acf{mpe} improvement over using only feature-engineered domain-specific features.

From a detailed per-feature analysis, we learned that besides dominant domain-specific features, behavioral characteristics, such as ratio or consumption a day and week ago, and contextual data related to the part of the day and sun position are the most significant contributors to the model's accuracy.

Furthermore, we demonstrated the importance of sub-metering data. The comparison shows up to 13\% \ac{mse} and 3\% \ac{mpe} loss of forecasting accuracy when not using sub-meters. However, when sub-meter data is not directly available, Non-Intrusive Load Monitoring or signal disaggregation techniques can be used to estimate appliance-level energy consumption from high-frequency aggregate power data. \ac{nilm} and signal disaggregation can be used to estimate sub-metered data, allowing for accurate feature engineering even in situations where sub-meters are not available or feasible to install.

Investing time and resources into further research on \ac{nilm} and signal disaggregation techniques is worthy, as they offer an alternative to sub-metering and can improve forecasting accuracy in situations where sub-meter data is not directly available. Future work can focus on developing more accurate and efficient techniques, as well as exploring their applicability in different settings, such as commercial buildings or industrial facilities.

%These techniques provide detailed information about individual appliances' energy usage, which can be used to extract informative features for forecasting models, improving their accuracy. Additionally, NILM and signal disaggregation can be used to estimate sub-metered data, allowing for accurate feature engineering even in situations where sub-meters are not available or feasible to install

% Using a consumption forecasting example, we show that 1) contextual features are found to be almost as important and domain specific features and 2) by adding additional contextual and behavioral aspects to the typical feature set decreases the prediction error (\ie, mean average error) by 11\% from 0.308 kWh to 0.274 kWh.

\section*{Acknowledgment}
This work was funded by the Slovenian Research Agency under the Grant P2-0016 and the European Commission under grant number 872613. We would also like to acknowledge Anže~Pirnat for his help with collecting some of the metric names and meta-data for the taxonomy. 




%\bibliographystyle{IEEEtran}
\bibliographystyle{elsarticle-num}
\bibliography{biblio}

\end{document}